\documentclass{article}

\usepackage{arxiv}
\pdfoutput=1
\usepackage[numbers]{natbib}
\usepackage[utf8]{inputenc} 
\usepackage[T1]{fontenc}    
\usepackage{hyperref}       
\usepackage{url}            
\usepackage{booktabs}       
\usepackage{amsfonts}       
\usepackage{nicefrac}       
\usepackage{microtype}      
\usepackage{lipsum}
\usepackage{graphicx}
\graphicspath{ {./images/} }
\usepackage{amssymb}
\usepackage{amsmath,amssymb,amsfonts}
\usepackage{algorithm}

\usepackage{graphicx}
\usepackage{booktabs}
\usepackage{multicol}
\usepackage{hyperref}
\usepackage{tabularx}
\usepackage{natbib} 
\usepackage{graphicx}  
\usepackage{glossaries}
\usepackage{algpseudocode} 
\usepackage{amsmath}  
\usepackage{algorithm}
\usepackage{algpseudocode}

\usepackage{booktabs}

\title{Optimizing  4th-Order Runge-Kutta Methods: A Dynamic Heuristic Approach for Efficiency and Low Storage}

\author{
 Gavin Lee Goodship  \\
  School of Computer Science\\
  TU Dublin \\
  Grangegorman Lower, Dublin 7, D07 H6K8, Ireland \\
  \texttt{D23126880@mytudublin.ie} \\
   \And
 Luis Miralles-Pechuan \\
  School of Computer Science \\
  TU Dublin \\
  Grangegorman Lower, Dublin 7, D07 H6K8, Ireland\\
  \texttt{luis.miralles@tudublin.ie} \\
  \And
 Stephen
O’Sullivan \\
  School of Computer Science\
  TU Dublin\\
  Grangegorman Lower, Dublin 7, D07 H6K8, Ireland \\
  \texttt{stephen.osullivan@tudublin.ie} \\
}

\begin{document}
\maketitle
\begin{abstract}
Extended Stability Runge-Kutta (ESRK) methods are crucial for solving large-scale computational problems in science and engineering, including weather forecasting, aerodynamic analysis, and complex biological modelling. However, balancing accuracy, stability, and computational efficiency remains challenging, particularly for high-order, low-storage schemes. This study introduces a hybrid Genetic Algorithm (GA) and Reinforcement Learning (RL) approach for automated heuristic discovery, optimizing low-storage ESRK methods. Unlike traditional approaches that rely on manually designed heuristics or exhaustive numerical searches, our method leverages GA-driven mutations for search-space exploration and an RL-inspired state transition mechanism to refine heuristic selection dynamically. This enables systematic parameter reduction, preserving fourth-order accuracy while significantly improving computational efficiency.

The proposed GA-RL heuristic optimization framework is validated through rigorous testing on benchmark problems, including the 1D and 2D Brusselator systems and the steady-state Navier-Stokes equations. The best-performing heuristic achieves a 25\% reduction in IPOPT runtime compared to traditional ESRK optimization processes while maintaining numerical stability and accuracy. These findings demonstrate the potential of adaptive heuristic discovery to improve resource efficiency in high-fidelity simulations and broaden the applicability of low-storage Runge-Kutta methods in real-world computational fluid dynamics, physics simulations, and other demanding fields. This work establishes a new paradigm in heuristic optimization for numerical methods, opening pathways for further exploration using Deep RL and AutoML-based heuristic search
\end{abstract}


\section{Introduction}
\label{sec:intro}
Extended Stability Runge-Kutta (ESRK) schemes of the fourth order are essential for solving mildly stiff ordinary and partial differential equations (ODEs and PDEs). These methods are widely applied in computational fluid dynamics, climate modeling, and various engineering and physics simulations, where high accuracy and numerical stability are crucial \cite{Ibrahim2010A}. ESRK schemes effectively balance computational efficiency and stability, making them ideal for handling fine-scale physical phenomena such as turbulence modeling in fluid dynamics and long-term trend prediction in climate systems \cite{Borodulin2021Explicit}.

Despite their effectiveness, existing ESRK schemes face fundamental challenges in handling stiff equations efficiently. The primary difficulty lies in the high computational costs of finding optimal coefficients that ensure extended stability while maintaining 4th-order accuracy \cite{Alamri2022Very}. As the number of stages in a scheme increases, the complexity of optimising the coefficients grows exponentially, often leading to impractically long optimisation times and excessive computational overhead \cite{Boom2018Optimization}. This limitation restricts the scalability of ESRK schemes for large-scale scientific simulations.

A significant issue in ESRK optimisation is the large number of free parameters in the Butcher tableau, which define the coefficients of the scheme. Standard optimisation approaches, such as Interior Point Optimisers (IPOPT), must search a vast parameter space to find suitable coefficients, making the process computationally expensive. Without constraints, the optimiser explores an unnecessarily large search space, leading to longer runtimes and higher memory usage. To address this, reducing the number of free parameters is crucial for improving the efficiency and feasibility of ESRK methods \cite{Alamri2022Very}.

This work introduces a heuristic-driven approach to reduce the number of free parameters in ESRK schemes systematically. We impose structural constraints on the coefficient optimization process by applying intelligent heuristics, limiting the search space. The dimensionality reduction significantly reduces computational costs because fewer parameters must be optimized. Rather than relying on manually derived heuristics, which can be restrictive and time-consuming, we propose an automated heuristic discovery framework that combines Genetic Algorithms (GA) and Reinforcement Learning (RL). Through iterative feedback loops, our method dynamically refines heuristics that improve computational efficiency while preserving numerical accuracy.

The reduction of free parameters has several key benefits. First, it lowers computational costs by reducing the time required for numerical optimization. Second, it improves scalability by making ESRK schemes more practical for large-scale simulations, such as climate modeling and aerodynamics. Third, it enhances memory efficiency by structuring the Butcher tableau to require less storage, making it particularly beneficial for high-performance computing applications. Additionally, these improvements expand the applicability of ESRK methods to various fields, including computational fluid dynamics, real-time control systems, and physics-based simulations, where fast and stable numerical integration is essential.

Our paper presents an automated heuristic discovery framework for optimizing 4th-order, 16-stage low-storage Extended Stability Runge-Kutta (ESRK) schemes. Rather than directly optimizing individual coefficients, this work optimizes the coefficient selection process by identifying structural relationships within the Butcher tableau. By reducing the number of free parameters beyond conventional low-storage limits, our approach enhances computational efficiency while preserving stability and 4th-order accuracy.

By leveraging Genetic Algorithms (GA) and Reinforcement Learning (RL), we dynamically generate and refine heuristics that systematically reduce the search space. This guides coefficient selection to enforce structural constraints while maintaining numerical accuracy. The best heuristics discovered lead to a 25\% reduction in computational runtime compared to traditional optimization methods, significantly enhancing efficiency.

We validate the effectiveness of our approach through rigorous testing on benchmark problems, including the 1D and 2D Brusselator systems and the Navier-Stokes equations. Our results demonstrate that the reduced-parameter schemes maintain 4th-order accuracy, stability, and a low-storage structure.

This study sets a new standard for optimizing ESRK schemes by integrating heuristics to minimize the parameter space. The remainder of this paper is structured as follows: Section \ref{section:literature} reviews relevant literature on ESRK schemes and heuristic optimization techniques. Section \ref{section:method} describes the methodology, including integrating GAs and RL with IPOPT for heuristic development. Section \ref{section:experiments} presents experiments validating the proposed framework, while Section \ref{section:conclusions} compares the results with existing approaches and outlines potential directions for future research.s.

\section{Literature Review}\label{section:literature}
This section overviews previous work on extended stability Runge-kutta schemes, commonly referred to as (ESRK). The low storage implementation of Runge-kutta schemes, commonly referred to as (LSRK), Interior point optimizers (IPOPT) benefits and drawbacks, and the Reinforcement learning (RL) and Genetic Algorithm (GA) operators make the explorations and exploitation of the search space possible.

\subsection{Extended stability Runge Kutta Schemes}
Fourth-order Extended Stability Runge-Kutta (ESRK) schemes are widely employed in solving ordinary differential equations (ODEs) and partial differential equations (PDEs), particularly for stiff problems. These methods extend the stability region along the negative real axis, making them suitable for problems with significant eigenvalue disparities, such as reaction-diffusion systems and climate modelling.

The development of 4th-order ESRK schemes has focused on balancing high-order accuracy with extended stability. Carpenter and Kennedy \cite{carpenter1994low} introduced a family of 4th-order low-storage Runge-Kutta schemes designed to reduce memory usage while maintaining stability for hyperbolic and parabolic problems. Their work demonstrated that 4th-order methods could achieve extended stability without sacrificing computational efficiency, laying the foundation for many modern ESRK approaches.

The ESRK framework was further extended with the introduction of Factorized Runge-Kutta-Chebyshev (FRKC) methods, which utilize Chebyshev polynomials to optimize stability properties for explicit methods \cite{osullivan2017factorized}. This advancement broadened the applicability of ESRK schemes to stiff PDEs, particularly in diffusive systems, while effectively reducing numerical dissipation and dispersion errors.

Another significant contribution is the work by Berland et al. \cite{berland2006low}, which focused on low-dissipation and low-dispersion 4th-order schemes for acoustics and wave propagation. Their schemes optimize stability and accuracy by minimizing numerical errors over the desired stability domain.

Current challenges in 4th-order ESRK schemes include balancing constraints, reducing computational overhead, and expanding stability regions without compromising accuracy. These challenges are particularly relevant in low-storage implementations, where the number of registers is minimized to enhance computational efficiency. This work's novel contributions aim to address these issues, extending the applicability of 4th-order ESRK schemes to more demanding computational problems.

\begin{table}[h]
    \centering
    \caption{Advantages and Disadvantages of Low Storage Schemes}
    \label{tab:lsrk_adv_disadv}
    \begin{tabular}{|p{4cm}|p{8cm}|}
        \hline
        \textbf{Category} & \textbf{Description} \\ 
        \hline
        \multicolumn{2}{|c|}{\textbf{Advantages of Low Storage Schemes}} \\
        \hline
        \textbf{Memory Efficiency} & LSRK schemes reduce memory footprint by storing only a limited number of intermediate solution vectors, which is crucial for large-scale systems and high-dimensional problems \cite{kennedy2000low}. \\
        \hline
        \textbf{Scalability} & Well-suited for large-scale simulations on parallel computing architectures. The reduced memory usage enhances scalability on distributed systems and GPUs \cite{carpenter1994fourth, wang2016low}. \\
        \hline
        \textbf{Computational Performance} & Reducing storage requirements leads to fewer memory accesses, improving computational speed, particularly on memory-bound systems. Ideal for high-performance computing applications \cite{toulorge2012optimal}. \\
        \hline
        \multicolumn{2}{|c|}{\textbf{Disadvantages of Low Storage Schemes}} \\
        \hline
        \textbf{Complex Coefficient Structures} & Achieving high-order accuracy and extended stability requires intricate coefficient structures, complicating implementation and analysis. Coefficient selection often demands advanced optimization techniques \cite{kennedy2000low, toulorge2012optimal}. \\
        \hline
        \textbf{Limited to Explicit Schemes} & Typically designed for explicit methods, making them less effective for stiff systems that require implicit approaches \cite{wang2016low}. \\
        \hline
        \textbf{Reduced Flexibility in Intermediate Stages} & The low storage constraint limits the number of intermediate solutions that can be stored, making it difficult to optimize stability properties or match the performance of high-storage schemes in certain cases \cite{kennedy2000low}. \\
        \hline
    \end{tabular}
\end{table}

Low-storage schemes strike a balance between \textit{memory efficiency} and \textit{computational performance}, making them highly beneficial for large-scale simulations. Despite challenges such as complex coefficient optimization and constraints in handling stiff problems, their scalability and reduced resource demands make them an essential approach for modern scientific and engineering computations. As summarized in Table \ref{tab:lsrk_adv_disadv}, these schemes offer significant advantages in terms of efficiency and performance while presenting trade-offs that must be carefully considered depending on the application.

\subsubsection{Construction of ESRK Schemes using an Interior point optimizer}
Developing a 4th-order sixteen-stage ESRK scheme requires solving an optimization problem to determine the appropriate Butcher tableau coefficients. An Interior Point Optimizer (IPOPT) ensures the resulting scheme satisfies stability and order condition requirements. Proper optimization problem formulation is essential to maintain well-posedness and computational feasibility.

One of the main challenges in constructing ESRK schemes using an interior point optimizer (IPOPT) is that the search space has to be well-defined. If a narrow search space is defined, the IPOPT will struggle to find solutions, leading to convergence issues. If the search space is too large, long compute times could occur.

As outlined by Biegler and Zavala~\cite{Biegler2009Large-scale}, careful tuning of the search space is essential for large-scale nonlinear programming problems. In this work, the search space is defined as:  
\[
\text{Search Space} \sim U(0, \, \frac{1}{s}),\label{eq:search}
\]
Where \( s \) represents the number of stages in the ESRK scheme. This specific choice of search space has proven effective for the sixteen-stage ESRK scheme, balancing computational efficiency with the feasibility of solutions.

\subsection{Genetic Algorithms (GA)}
Genetic Algorithms (GA) are a class of evolutionary optimization techniques inspired by natural selection. They are widely used in numerical analysis, particularly for optimizing coefficients in Runge-Kutta schemes, where traditional gradient-based methods struggle with complex and non-convex search spaces. As summarized in Table \ref{tab:ga_esrk}, GA offers significant advantages, such as global search capabilities and robustness to non-smooth problems, making it a powerful tool for Extended Stability Runge-Kutta (ESRK) optimization. However, challenges like high computational cost and the need for careful search space design must be considered when applying GA to ESRK coefficient optimization.

\begin{table}[h]
    \centering
    \caption{Advantages and Challenges of Genetic Algorithms in ESRK Optimization}
    \label{tab:ga_esrk}
    \begin{tabular}{|p{4cm}|p{8cm}|}
        \hline
        \textbf{Category} & \textbf{Description} \\ 
        \hline
        \multicolumn{2}{|c|}{\textbf{Advantages of GA in ESRK Optimization}} \\
        \hline
        \textbf{Global Search Capabilities} & GA effectively explores large and complex search spaces by maintaining a population of candidate solutions. Through selection, crossover, and mutation, it refines solutions, allowing it to escape local minima and identify near-optimal coefficients for ESRK schemes. \\
        \hline
        \textbf{Robustness to Non-Smooth Problems} & Unlike gradient-based methods, GA does not require continuity or differentiability of the objective function. This makes it particularly effective for optimizing stability regions and accuracy in Runge-Kutta schemes. \\
        \hline
        \multicolumn{2}{|c|}{\textbf{Challenges of GA in ESRK Optimization}} \\
        \hline
        \textbf{Computational Cost} & GA often requires many iterations and fitness evaluations to achieve convergence, leading to high computational costs, especially for high-dimensional optimization problems. \\
        \hline
        \textbf{Search Space Design} & Defining a suitable search space is critical for GA’s success. A poorly designed search space can significantly increase optimization time or lead to infeasible solutions. \\
        \hline
    \end{tabular}
\end{table}

\subsection{Reinforcement Learning (RL)}

Reinforcement Learning (RL) has gained recognition as a powerful optimization framework, where an agent iteratively improves solutions through interactions with its environment. In the context of ESRK coefficient optimization, RL provides a dynamic, data-driven method for efficiently exploring the search space to discover heuristics that meet strict order conditions and stability requirements. As summarized in Table \ref{tab:rl_esrk}, RL offers advantages such as adaptive learning and self-improvement, making it well-suited for optimizing multi-stage, high-order numerical schemes. However, challenges like reward function design, computational overhead, and the exploration-exploitation tradeoff must be carefully managed to ensure effective optimization.

\begin{table}[h]
    \centering
    \caption{Advantages and Challenges of Reinforcement Learning in ESRK Optimization}
    \label{tab:rl_esrk}
    \begin{tabular}{|p{4cm}|p{8cm}|}
        \hline
        \textbf{Category} & \textbf{Description} \\ 
        \hline
        \multicolumn{2}{|c|}{\textbf{Advantages of RL in ESRK Optimization}} \\
        \hline
        \textbf{Adaptive Learning} & RL algorithms dynamically adjust their strategies based on feedback (rewards) from the environment. This enables efficient exploration and exploitation of the search space to find suitable ESRK coefficients. \\
        \hline
        \textbf{Self-Improvement} & RL systems improve their performance over time as they accumulate experience through interactions. This iterative learning process is particularly beneficial for optimizing multi-stage, high-order numerical schemes. \\
        \hline
        \multicolumn{2}{|c|}{\textbf{Challenges of RL in ESRK Optimization}} \\
        \hline
        \textbf{Reward Function Design} & The success of RL depends heavily on how the reward function is defined. It must effectively balance multiple objectives, such as accuracy, stability, and computational cost, to guide the optimization process efficiently. \\
        \hline
        \textbf{Computational Overhead} & RL algorithms often require numerous training episodes to converge to an optimal policy. This leads to significant computational overhead compared to traditional optimization methods. \\
        \hline
        \textbf{Exploration vs. Exploitation Dilemma} & RL must balance exploration (searching for new solutions) with exploitation (refining known good solutions). Improper balancing can result in suboptimal performance or inefficient convergence. \\
        \hline
    \end{tabular}
\end{table}
\section{Methodology}\label{section:method}
This section details the approach for optimizing the coefficient structures of the Butcher tableaus with heuristics in the 4th-order sixteen-stage scheme. This section starts with implementing the reduced scheme constraints, in which the methodology finds heuristics to minimize the iterations that the interior point optimizer takes to find convergence solutions. All heuristics are found on the Van der Houwen reduced scheme \cite{carpenter1994low}.

Our methodology combines elements of Genetic Algorithms (GA) and implicit Reinforcement Learning (RL) for heuristic optimization. In this framework, we define a 'state' based on whether a heuristic leads to a stable ESRK scheme that meets order conditions. If a heuristic is successful, it is stored for further refinement (exploitation). If it fails, we apply genetic mutations to explore new heuristics (exploration). This iterative refinement mimics policy learning in RL, where mutation acts as the primary mechanism for policy improvement. However, unlike traditional RL, our approach does not rely on an explicit reward function but instead leverages binary feedback (stable vs. unstable) to drive heuristic evolution.

\begin{figure}[h]
    \centering
    \includegraphics[scale=0.20]{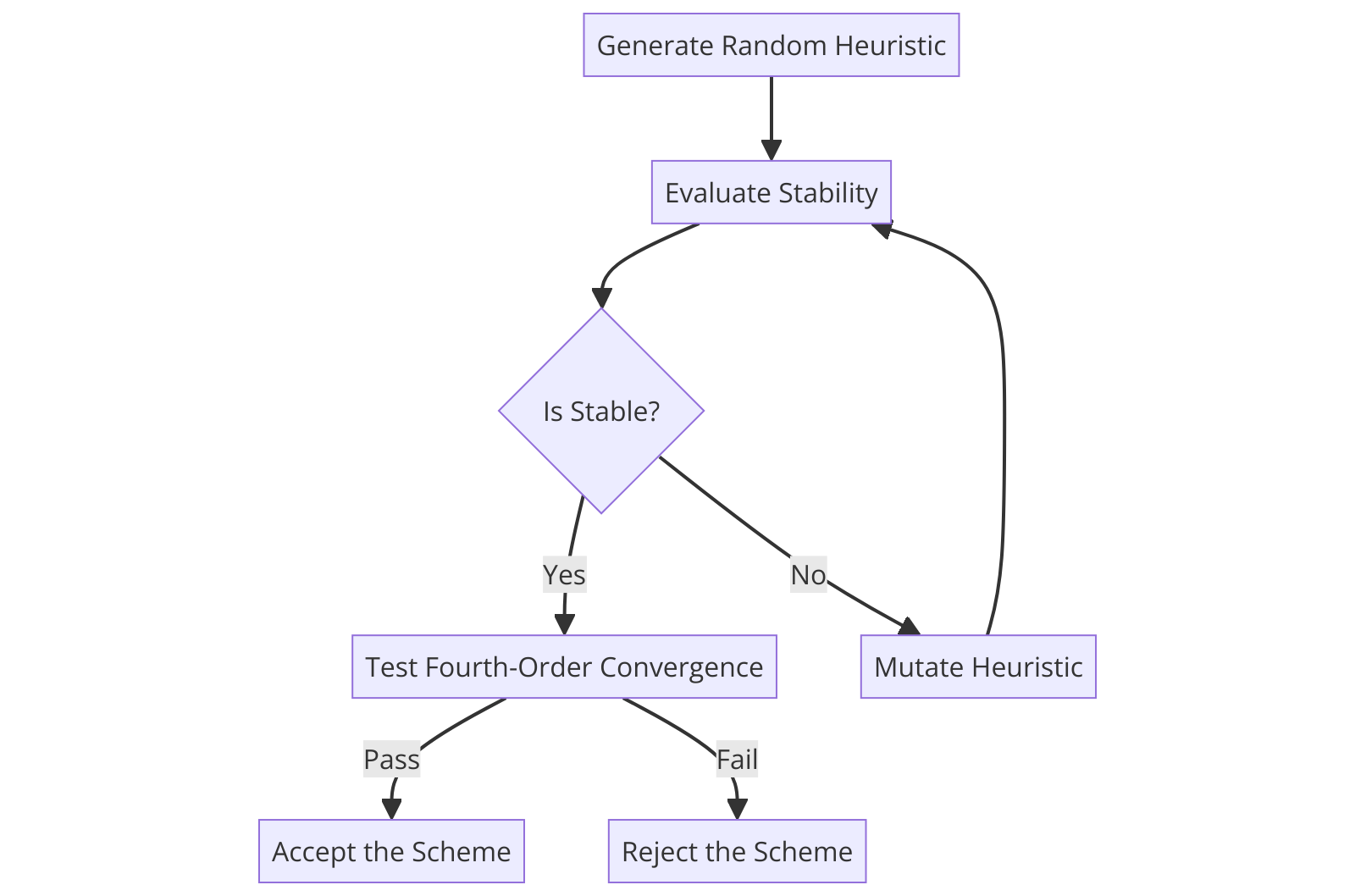}
    \caption{An overview of the fourth order reduced scheme methodology pipeline. This pipeline overviews the methodology, from mutating the expressions to order convergence studies.}
    \label{fig:methodology}
\end{figure}

Even with the reduced scheme, heuristics can be found, which reduces the number of iterations to find a convergent solution without any heuristics added to the Butcher tableau; the average run time over 100 runs has a mean run time of 2010 iterations. In this methodology, to find heuristics, we use genetic operators to explore the search space, and the constraints function as exploit action where if the heuristics produce the correct stability polynomial, the heuristics are then stored for further analysis. If the stability polynomial check fails, the heuristic is then mutated through genetic operators. Figure \ref{fig:methodology} shows an overview of the methodology pipeline.

The below pipeline table \ref{tab:pipeline} shows our systematic approach to optimize the ESRK scheme. Starting from a constrained initial state, random heuristics are generated and refined through RL feedback and GA mutations. Stability and convergence analyses are conducted at each step to ensure that the final scheme meets both efficiency and accuracy requirements. This structured methodology allows for robust performance improvements, even in an already optimized, low-storage framework

\begin{table}[h!]
\centering
\caption{Overview of the Methodology Pipeline}
\label{tab:pipeline}
\begin{tabular}{|l|p{8cm}|}
\hline
\textbf{Step} & \textbf{Description} \\
\hline
Initialization & Start the IPOPT without heuristic values using the constrained tableau structure. \\
\hline
Random Heuristic Generation & Generate a random heuristic and integrate it into the order conditions function. \\
\hline
Constraint Application & Apply order condition constraints to the coefficients using the heuristic. \\
\hline
Stability Validation & Compute the stability polynomial and validate the scheme's stability. \\
\hline
Heuristic Refinement & Use RL feedback loops to refine heuristics; if a heuristic fails, apply mutations up to a predefined limit. \\
\hline
Fallback Mechanism & Generate new heuristics if the mutation limit is reached without success. \\
\hline
Convergence Analysis & Perform convergence analysis to confirm the correct order of convergence. \\
\hline
\end{tabular}
\end{table}

\subsubsection*{Reduced scheme for the parameter space reduction}

To reduce the free parameters in the Runge-Kutta tableau, we constrain most coefficients in the $A$-matrix to the corresponding $b$-values. This approach minimises storage requirements while maintaining computational efficiency. The A matrix constraints are implemented with

\label{alg:apply_constraints}
\begin{algorithm}
\caption{Apply Constraints}
\begin{algorithmic}[1]
\For{$j \gets 0$ \textbf{to} $s{-}1$}
  \For{$i \gets j{+}1$ \textbf{to} $s{-}1$}
    \If{$(i,j) \in \mathcal{F}$}
      \State $a_{i,j} \gets \text{free parameter}$
    \Else
      \State $a_{i,j} \gets b_j$
    \EndIf
  \EndFor
\EndFor
\end{algorithmic}
\end{algorithm}

For a 16-stage, 4th-order scheme, the free indices are:
\[
\text{Free Indices} = \{(1,0),(2,1), (3,2), (4,3), \dots, (16,15)\}.
\]

The remaining $A$-matrix coefficients are constrained to their corresponding $b$-values. This structure ensures reduced storage while preserving stability and accuracy. It is commonly known as the Van der Houwen structure, where only 2S-1 unique coefficients are needed to produce a valid scheme \cite{Kennedy1999} where S is the number of stages in the Butcher tableau. An overview of the reduced is as follows, referenced in equation \ref{eq:reduced_tableau}.

\begin{equation}\label{eq:reduced_tableau}
\begin{array}{c|cccccc}
 0  &   &    &   &   &   &      \\
 c_1 & a_{0}  &    &   &   &   &      \\
 c_2 & b_{0}  & a_{1}   &   &   &   &      \\
 c_3 & b_{0}  & b_{1}   & a_{2}   &   &   &     \\
\vdots & \vdots & \vdots  & \vdots   & \ddots  &   &      \\
 c_{s-1} & b_{0}  & b_{1}   & b_{2}   & \cdots  & a_{s-2}  &      \\
\hline
    & b_0  & b_1   & b_2   & \cdots  & b_{s-2}  & b_{s-1}     
\end{array}
\end{equation}

\subsection{Order Condition Constraints Function Design}
A constraints function generates a 16-stage 4th-order scheme that ensures the 4th-order conditions are met. For a 4th-order Runge-Kutta scheme, the coefficients \(a_{ij}\), \(b_i\), and \(c_i\) must satisfy the Butcher order conditions.

\[
\begin{aligned}
\text{1st-order:} \quad &\sum_{i=1}^{16} b_i = 1, \\
\text{2nd-order:} \quad &\sum_{i=1}^{16} b_i c_i = \frac{1}{2}, \\
\text{3rd-order:} \quad &\sum_{i=1}^{16} b_i c_i^2 = \frac{1}{3}, \\
&\sum_{i=1}^{16} b_i \sum_{j=1}^i a_{ij} c_j = \frac{1}{6}, \\
\text{4th-order:} \quad &\sum_{i=1}^{16} b_i c_i^3 = \frac{1}{4}, \\
&\sum_{i=1}^{16} b_i \Big( \sum_{j=1}^i a_{ij} c_j \Big) c_i = \frac{1}{8}, \\
&\sum_{i=1}^{16} b_i \sum_{j=1}^i a_{ij} c_j^2 = \frac{1}{12}, \\
&\sum_{i=1}^{16} b_i \sum_{j=1}^i \sum_{k=1}^j a_{ij} a_{jk} c_k = \frac{1}{24}.
\end{aligned}
\]

These conditions impose constraints on the coefficients in the Butcher tableau, which consists of 120 \(a\)-values and 16 \(b\)-values for a 16-stage scheme. The constraints function ensures that all these equations are satisfied, generating a valid 4th-order scheme.

Once the Butcher tableau is determined, it is passed to the stability constraint evaluation function to verify if the heuristic produces a stable solution. This dual check, which ensures order conditions and stability, guarantees that the optimized scheme is accurate and robust for stiff differential equations.

\textbf{Stability Constraint Design:} Heuristics are guided toward stable solutions using the constraints function, which computes the stability polynomial.The stability polynomial \(R(z)\) for an \(s\)-stage Runge-Kutta scheme is given by
\[
R(z) = 1 + zb^T(I - zA)^{-1}\mathbf{1} = \sum_{j=0}^s \beta_j z^j,
\]
where \(z = \lambda \Delta t\) and the coefficients \(\beta_j\) must satisfy the order conditions for 4th-order accuracy:
\[
\beta_j = \frac{1}{j!} \quad \text{for } j = 0,\, 1,\, 2,\, 3,\, 4.
\]
This ensures that \(R(z)\) approximates the exponential \(e^z\) up to the fourth order. Moreover, for stability, the polynomial must remain bounded within the desired stability domain:
\[
|R(z)| \leq 1 \quad \text{for } z \text{ in the stability region,}
\]
Particularly along the negative real axis where stiff problems demand extended stability.

After the Butcher tableaus are calculated, the number of iterations is checked against the average (2010). The stability evaluation function follows these steps:
\begin{enumerate}
    \item Compute the stability polynomial \(R(z)\) using the current coefficients.
    \item Verify that the coefficients satisfy \(\beta_j = \frac{1}{j!}\) for \(j \leq 4\).
    \item Check that \(|R(z)| \leq 1\) within the targeted stability domain.
    \item Compare the number of iterations to the average benchmark (2010) to assess heuristic efficiency.
\end{enumerate}

\setcounter{enumi}{4} 

\begin{figure}[h]
    \centering
    \includegraphics[scale=0.30]{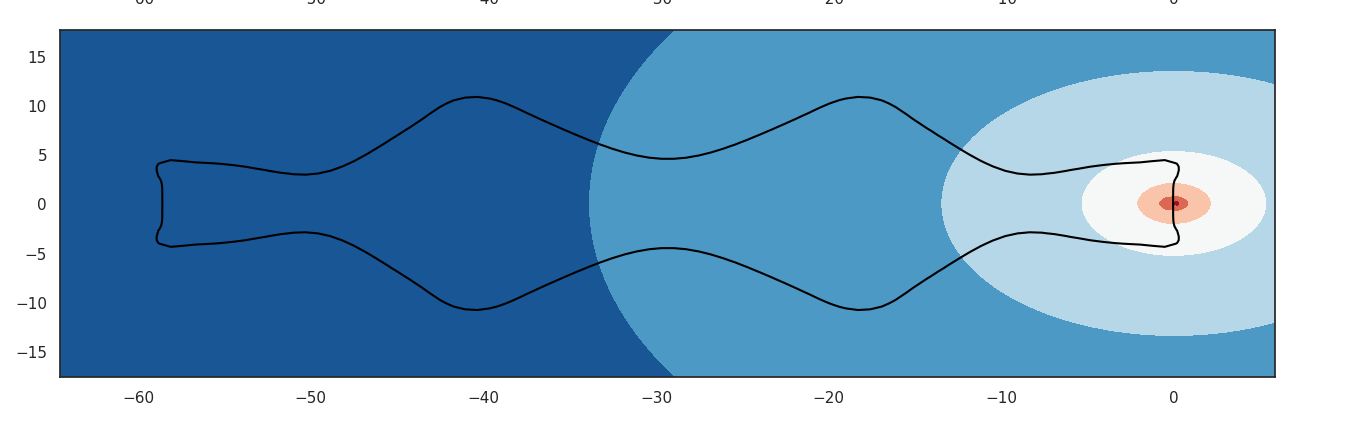}
\caption{Stability polynomial generated from the IPOPT. The color gradient represents the magnitude of the stability function, with darker blue indicating lower values and lighter colors (up to red) indicating higher values. Black contour lines represent the stability region boundaries.}
    \label{fig:fourth_order_poly}
\end{figure}

\begin{enumerate}
    \item \textbf{Initialization of Symbols}: Initialize symbolic variables for accurate and complex calculations to define the stability function in the complex plane.
    \item \textbf{Matrix Construction}: Convert the list of coefficients (\texttt{a\_values}) into a SymPy matrix, reshape it into a lower triangular matrix, and move the last row to the top for specific stability evaluations.
    \item \textbf{Definition of Coefficient Functions}: Define symbolic functions for weight coefficients (\texttt{b\_values}) and corresponding vectors and matrices, which are crucial for stability analysis.
    \item \textbf{Calculation of Stability Matrix}: Compute the stability matrix \( B(z) \) as:
    \[
    B(z) = I - zA
    \]
    where \( I \) is the identity matrix, \( z \) is a complex variable, and \( A \) is the coefficient matrix. The inverse of this matrix is used to calculate the stability function \( R(z) \).
    \item \textbf{Stability Function Evaluation}: Calculate the stability function \( R(z) \) as:
    \[
    R(z) = 1 + z b^T (I - zA)^{-1} \mathbf{1}
    \]
    Simplify and evaluate this function over a grid of complex values to capture the stability region.
    \item \textbf{Validation of Schemes}: For a scheme to be valid, the sum of roots of the stability polynomial is computed for the 4th-order stability polynomial (see Figure \ref{fig:fourth_order_poly}).
\end{enumerate}

\begin{figure}[h]
    \centering
    \includegraphics[scale=0.30]{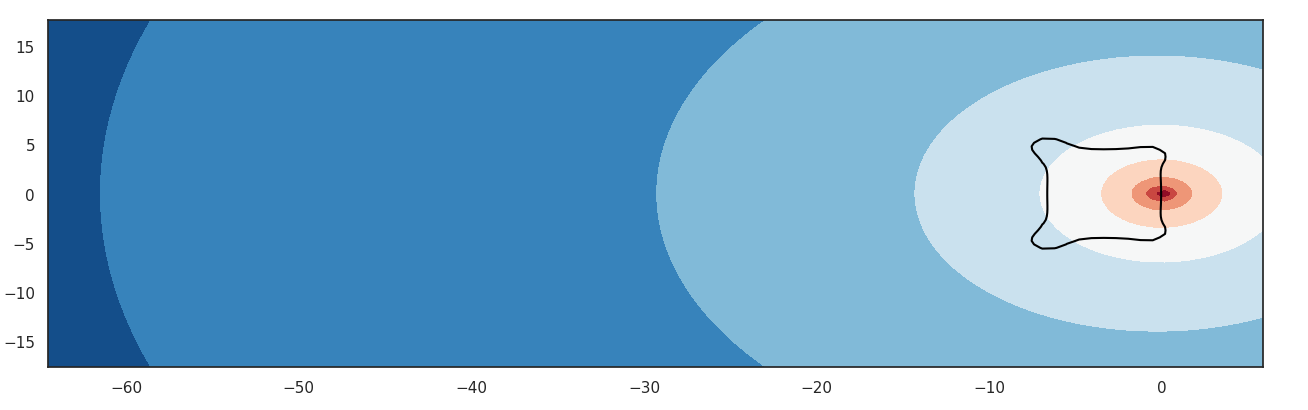}
\caption{Unstable Stability polynomial generated from the IPOPT. The colour gradient represents the magnitude of the stability function, with darker blue indicating lower values and lighter colours (up to red) indicating higher values. Black contour lines represent the stability region boundaries.}
    \label{fig:unstable}
\end{figure}

If the validation fails through the failure stability criteria, the random expression is then mutated up to a specific limit when a heuristic fails the stability constraint check; the polynomial can shrink towards zero on the real axis. This can be seen in graph \ref{fig:unstable}. Once the mutation starts, the stability is checked on every expression mutation until either the correct stability polynomial \ref{fig:fourth_order_poly} is found or a new expression is generated through the fullback mechanism.

The Butcher Tableaus coefficients are created through an interior point optimizer, in which random mutation and complete back mechanisms interact to receive and obtain the current state of the stability polynomial behavior. The settings have to be carefully crafted to balance precision and run time. The following settings are used with the methodology.

\subsection*{IPOPT Solver Settings}

The optimisation solver IPOPT was configured with the following settings to ensure precise and stable numerical solutions. These configurations, as outlined in Table \ref{tab:solver_config}, were carefully selected to enhance convergence robustness and maintain numerical stability when solving large-scale optimisation problems. By enforcing strict variable bounds while minimising constraint violations, these settings contribute to the reliability of the solver in handling complex numerical challenges.

A key component of this configuration is the use of the MUMPS linear solver, which is particularly efficient for solving large-scale sparse systems. This choice optimises computational performance by reducing overhead while maintaining solution accuracy. Additionally, the IPOPT solver version 3.13 was employed in this methodology, ensuring compatibility with modern numerical optimisation techniques.

To achieve a high level of numerical precision, the Bettol tolerance level was set to \(10^{-14}\). This strict tolerance ensures that the reduced scheme structure adheres to a well-defined numerical pattern, preventing deviations that could lead to stability issues. This level of precision is especially critical for applications such as low-storage numerical methods and stiff system solvers, where stable and accurate integration techniques are required.

Overall, these solver settings were specifically tailored for high-precision optimisation tasks, particularly in low-storage Runge-Kutta methods and numerical solvers for stiff differential equations. The combination of advanced solver techniques and fine-tuned tolerances enables a balance between computational efficiency and numerical accuracy, making the optimisation process both scalable and reliable for real-world scientific computing applications.

\begin{table}[h]
    \centering
    \renewcommand{\arraystretch}{1.3} 
    \setlength{\tabcolsep}{6pt} 
    \caption{IPOPT Configuration Parameters and Adjustments}
    \label{tab:solver_config}
    \begin{tabular}{|l|l|p{6cm}|c|}
        \hline
        \textbf{Category} & \textbf{Parameter} & \textbf{Value / Adjustment} & \textbf{Default} \\
        \hline
        \multicolumn{4}{|c|}{\textbf{General Solver Configuration}} \\
        \hline
        & \texttt{max\_iter} & 500,000 iterations for complex optimization problems. & — \\
        & \texttt{nlp\_scaling\_method} & \texttt{'none'} to preserve variable and constraint scaling. & Enabled \\
        & \texttt{linear\_solver} & \texttt{'mumps'} for solving large-scale sparse systems. & — \\
        & \texttt{honor\_original\_bounds} & \texttt{'yes'} to enforce strict adherence to bounds. & \texttt{'no'} \\
        \hline
        \multicolumn{4}{|c|}{\textbf{Tightened Tolerance Parameters}} \\
        \hline
        & \texttt{tol} & \texttt{bettol}, stricter than default. & $10^{-8}$ \\
        & \texttt{dual\_inf\_tol} & $10^8 \times \texttt{bettol}$ to relax dual feasibility. & $1$ \\
        & \texttt{constr\_viol\_tol} & $10^4 \times \texttt{bettol}$ to reduce constraint violation. & $10^{-4}$ \\
        & \texttt{compl\_inf\_tol} & $10^4 \times \texttt{bettol}$ to tighten complementarity slackness. & $10^{-4}$ \\
        \hline
        \multicolumn{4}{|c|}{\textbf{Acceptable Convergence Criteria}} \\
        \hline
        & \texttt{acceptable\_tol} & \texttt{bettol} threshold for acceptable convergence. & $10^{-6}$ \\
        & \texttt{acceptable\_dual\_inf\_tol} & $10^8 \times \texttt{bettol}$ to adjust acceptable infeasibility. & $10^{10}$ \\
        & \texttt{acceptable\_constr\_viol\_tol} & $10^4 \times \texttt{bettol}$ to strengthen constraint tolerance. & $10^{-2}$ \\
        & \texttt{acceptable\_compl\_inf\_tol} & $10^4 \times \texttt{bettol}$ for complementarity convergence. & $10^{-2}$ \\
        \hline
        \multicolumn{4}{|c|}{\textbf{Additional Bound Relaxation}} \\
        \hline
        & \texttt{bound\_relax\_factor} & \texttt{bettol} to improve stability in bound-constrained optimization. & $10^{-8}$ \\
        \hline
    \end{tabular}
\end{table}

These solver settings, as outlined in Table \ref{tab:solver_config}, were carefully selected to ensure robust convergence and numerical stability in large-scale optimisation problems. By enforcing strict variable bounds while minimising constraint violations, these configurations enhance the reliability of the optimisation process. The parameters were specifically tuned to balance computational efficiency and numerical accuracy, making them well-suited for complex optimisation tasks.

A key factor in improving solver efficiency is the use of the MUMPS linear solver, which is optimised for handling large-scale sparse systems. This choice significantly reduces computational overhead while maintaining solution accuracy. Additionally, the IPOPT solver version 3.13 was used in this methodology, ensuring compatibility with the latest numerical optimisation techniques.

To further refine the numerical precision, the Bettol tolerance level was set to \(10^{-14}\). This extremely strict tolerance ensures that the reduced scheme structure adheres to a well-defined numerical pattern, preventing deviations that could compromise stability. This level of precision is particularly important for low-storage numerical methods, where efficient memory usage is critical, and for stiff system solvers that require highly stable integration techniques.

Overall, these configurations are specifically tailored to high-precision applications, including low-storage Runge-Kutta methods and numerical solvers for mildly stiff differential equations. The combination of advanced solver techniques and fine-tuned tolerances enables a balance between computational efficiency and numerical accuracy, making the optimisation process both reliable and scalable for real-world scientific computing challenges.

\textbf{Random expression Generator}
Due to the reduced scheme being only 16 unique $a_{1,0}$ to $a_{16,15}$ with the associated b values of $b_{0}$ to $b_{15}$. We don't need a full scheme like the previous research performed \cite{goodship2024dynamic}. The search space reduces to near linear scalability when the Van der Houwen scheme is used \cite{Benner1998}.

\begin{algorithm}[ht]
\caption{GenRandExpr (restricted to reduced scheme)}
\label{alg:GenRandExpr}
\begin{algorithmic}[1]

\State \textbf{Input:} $maxLen$, $maxBnds$, $zTable$
\State \textbf{Output:} $expr$
\State $expr \gets \{\}$                       
\State $len  \gets \mathrm{rand}(1,\;maxLen)$  

\For{$i \gets 1$ \textbf{to} $len$}
  \State $type \gets \mathrm{rand}(\{a,b\})$   \Comment{only $a[i,j]$ or $b[j]$}
  \If{$type = \texttt{a}$}
    \State $j \gets \mathrm{rand}(1,16)$       \Comment{row index for $a_{i,j}$}
    \State $k \gets \mathrm{rand}(0,\,j{-}1)$  \Comment{column index $< j$}
  \Else
    \State $j \gets \mathrm{rand}(0,15)$       \Comment{index for $b_j$}
  \EndIf
  \State $expr \gets expr \;\|\; (type, j, k)$ 
\EndFor

\If{$len = 1$}
  \State $expr[0].power \gets \mathrm{rand}(1,6)$
\EndIf

\Return $expr$

\end{algorithmic}
\end{algorithm}

\textbf{Random Expression Mutation}.
For the reduced scheme random mutation expressions, only the coefficients allowed in the reduced scheme can be used for the mutation. Adapting this algorithm to only allow for the reduced scheme coefficients keeps the integrity of the low storage structure intact. This adaptation is done by not allowing other coefficients to be used in the random mutation outside the reduced scheme structure. 

\begin{algorithm}[ht]
\caption{Mutate random expression}
\label{alg:MutateReducedHeuristicExpression}
\begin{algorithmic}[1]

\State \textbf{Input:} heuristic expression $expr$; reduced coefficient set $\{a[i,j],\, b[j]\}$
\State \textbf{Output:} mutated expression

\Procedure{MutateExpression}{$expr$}

  \State Choose a random mutation type from $\{\text{Replace},\text{Addition},\text{Deletion},\text{Combined}\}$

  \If{mutation type is \emph{Coefficient‐Replacement}}
      \State Replace a random coefficient in $expr$ by one from $\{a[i,j], b[j]\}$
  \ElsIf{mutation type is \emph{Term‐Addition}}
      \State Add a new term to $expr$ that uses only coefficients in $\{a[i,j], b[j]\}$
  \ElsIf{mutation type is \emph{Term‐Deletion}}
      \State Remove a random term from $expr$
  \ElsIf{mutation type is \emph{Combination}}
      \State Combine two random coefficients in $expr$ (add, multiply, or power) using only $\{a[i,j], b[j]\}$
  \EndIf

  \State \Return $expr$

\EndProcedure
\end{algorithmic}
\end{algorithm}

\textbf{Search Space Definition for Reduced Heuristic Generation:} To find Butcher tableaus that meet the necessary stability and accuracy using the reduced coefficient structure, the search space is defined as follows:

\begin{itemize}
    \item \textbf{Variables:} Only the available reduced coefficients $\{a[i,j]\}$ (unique under the diagonal) and $\{b[j]\}$ (shared or repeated) are used to form heuristic expressions.
    \item \textbf{Operations:} Permitted operations include addition, multiplication, and powers.
    \item \textbf{Maximum Expression Length:} Each expression has up to 3 terms.
    \item \textbf{Constraint:} Ensure all expressions align with the reduced structure, avoiding redundant or unused coefficients.
\end{itemize}

\subsection*{Example Mutation Process}

\noindent The following example demonstrates the mutation process:

\begin{enumerate}
    \item \textbf{Initial Failed Heuristic:}  
    Suppose the original heuristic expression is:  
    \[
    b_0 = a_{1,0} + b_2 + b_3
    \]  
    This heuristic is tested and fails the stability check during validation.

    \item \textbf{Mutation Process:}
    \begin{enumerate}
        \item \textbf{Step 1: Mutation Type Selection}  
        The algorithm selects a random mutation type, such as \textit{Term Addition}.

        \item \textbf{Step 2: Apply Mutation}  
        Add a new term to the heuristic using the reduced coefficient set:  
        \[
        b_0 = a_{1,0} + b_2 + b_3 + a_{2,1}
        \]

        \item \textbf{Step 3: Stability Check}  
        The modified heuristic is tested, but it still fails stability validation.

        \item \textbf{Step 4: Further Mutation}  
        Apply a second mutation, such as \textit{Coefficient Replacement}:  
        Replace \(a_{2,1}\) with \(b_4\), resulting in:  
        \[
        b_0 = a_{1,0} + b_2 + b_3 + b_4
        \]

        \item \textbf{Step 5: Re-validation}  
        Reapply the stability check. If the heuristic passes, the process stops.  
        If it fails, additional mutations are applied until the predefined limit is reached or the heuristic stabilises.
    \end{enumerate}

    \item \textbf{Result:}  
    After several iterations, the heuristic converges to a stable and valid form, such as:  
    \[
    b_0 = b_1 + b_2 + a_{3,2}
    \]  
    This form passes both the stability check and the order condition constraints.
\end{enumerate}

\textbf{Fallback mechanism}
If the heuristic fails over the predefined limit of the mutations, a new coefficient is found within the reduced scheme structure. It is picked randomly in this process, starting again with the mutations of the random expression generations. This methodology has been able to find heuristics that not only reduce the parameter space of the problem in a reduced scheme but also reduce the number of iterations that are needed to find a convergent solution. This fallback mechanism operates similarly to Monte Carlo Tree Search (MCTS) in reinforcement learning, where a search is restarted if a branch does not yield a promising solution. Additionally, the mutation process introduces genetic search principles, ensuring that heuristics evolve dynamically without requiring manually predefined search rules. This dual-layered approach—combining genetic exploration with RL-based state transitions—makes the optimisation framework highly adaptive and efficient.

\section{Results and Experiments}\label{section:experiments}

This section discusses the heuristics found on the reduced scheme and their overall effect on the interior point optimiser's runtime over 100 runs. All heuristics found in this study produced 4th-order convergence and the correct stability polynomial, and performing order convergence tests of each of the 100 heuristics' runs shows the robustness of the results.

\textbf{Baseline run time with no heuristics} The run over 100 runs of the IPOPT without heuristics produced the following runtime statistics.

\begin{table}[h!]
    \centering
    \begin{tabular}{|l|r|}
        \hline
        \textbf{Statistic}         & \textbf{Value} \\ \hline
        Count                      & 100            \\ 
        Mean                       & 2,010           \\ 
        Median                     & 792            \\ 
        Standard Deviation         & 5,987           \\ 
        Minimum                    & 291            \\ 
        Maximum                    & 55,671          \\ 
        \hline
    \end{tabular}
    \caption{Baseline statistics of the reduced 4th-order 16-stage ESRK scheme with  no heuristics.}
    \label{tab:baseline_stats}
\end{table}

The following heuristics and run times were found in the reduced scheme, with the following run times being over 100 runs of the IPOPT, and the subsequent data was collected.

\begin{table}[h!]
\centering
\small
\begin{tabularx}{\linewidth}{Xc}
\toprule
\textbf{Heuristic Expression} & \textbf{Iterations} \\
\midrule
$a_{5,4} = b_{8} + b_{13}$ & 1,802 \\
$b_{10} = a_{9,8} \cdot b_{9}$ & 1,917 \\
$b_{3} = b_{4} + a_{1,0} + a_{15,14}$ & 1,799 \\
$b_{5} = a_{2,1}^2$ & 1,501 \\
$b_{1} = a_{7,6} + b_{12}$ & 1,842 \\
\bottomrule
\end{tabularx}
\caption{Performance of low-storage scheme heuristics for the 4th-order, 16-stage ESRK method. Each heuristic expression is listed alongside the corresponding number of iterations required for the interior point optimiser to converge. These results highlight the impact of heuristic expressions on improving computational efficiency within the reduced parameter space framework.}
\label{table:Reduce_performance}
\end{table}

\subsubsection{Multiple Heuristics on the Reduced Scheme}  
Previous research typically employed a single heuristic at a time, producing meaningful results but leaving the question of how much further the parameter space could be reduced. This section demonstrates that multiple heuristics can be incorporated into the reduced scheme. By applying two heuristics, \( b_{5} = a_{2,1}^2 \) and \( a_{5,4} = b_{8} + b_{13} \), the parameter space is reduced further, consolidating the computational benefits of the reduced scheme while maintaining 4th-order accuracy and stability. The drawback of using multiple heuristics is that the runtime approaches the baseline runtime of 2010 iterations. In this example, over 100 iterations of the interior point optimiser (IPOPT) produced an average runtime of 1910 iterations. The parameter search space can be further reduced, but this comes at the cost of increased IPOPT runtime. However, if runtime is not a significant concern, the number of free parameters can be reduced from \(2S - 1\), demonstrating that there is still room for optimising the reduced Butcher tableau structure.

\textbf{Performance Analysis:}  
When two heuristics are added, the mean IPOPT runtime over 100 runs is reduced to 1,910 iterations, compared to the baseline runtime of 2,010 iterations for the reduced scheme without heuristics. Although this represents only a slight improvement in runtime, the substantial reduction in the tableau size highlights the efficiency of the reduced scheme in terms of storage and computation.

\begin{table}[h!]
\centering
\caption{Key Findings of the Study}
\label{tab:key_findings}
\begin{tabular}{|p{3cm}|p{8cm}|}
\hline
\textbf{Finding} & \textbf{Description} \\
\hline
Multiple Heuristics Integration & The integration of multiple heuristics into the reduced scheme is feasible and preserves the structural integrity of the low-storage framework. \\
\hline
Parameter Space Reduction & Dual heuristics significantly reduce the number of free parameters, thereby enhancing computational efficiency. \\
\hline
Runtime Improvement & Even though the scheme is already optimised with relatively few free coefficients, incorporating heuristics results in approximately a 25\% reduction in the iterations required for convergence, demonstrating a meaningful performance boost. \\
\hline
\end{tabular}
\end{table}

\begin{figure}[h]
    \centering
    \includegraphics[scale=0.50]{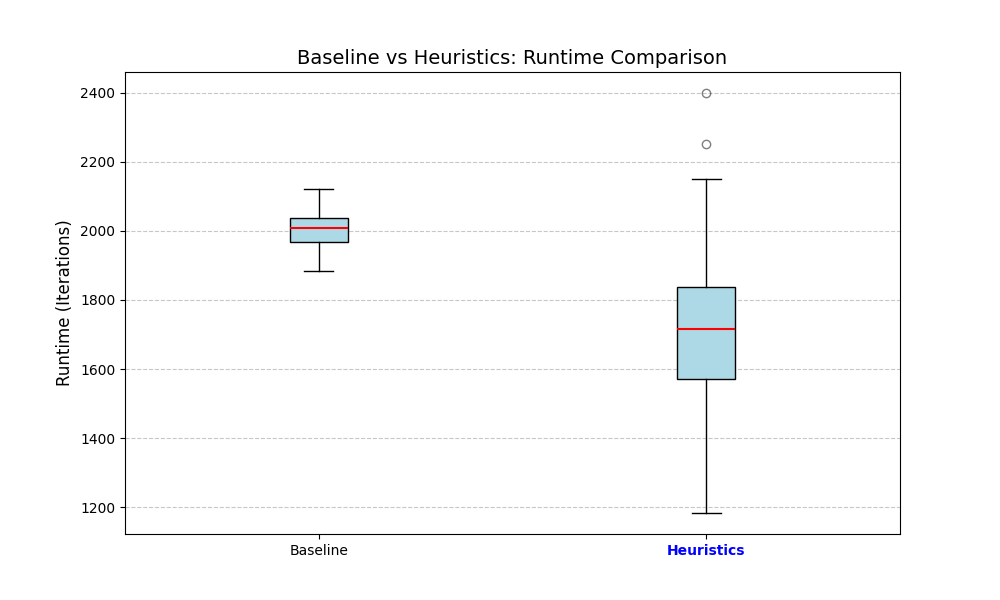}\label{fig:compare}
\caption{Baseline vs. heuristics box plot comparison of IPOPT run time with heuristics added vs. the baseline run times without any heuristics.}
\end{figure}

This work establishes a foundation for exploring the incorporation of additional heuristics into reduced schemes, offering a pathway to even more significant efficiency gains in future research. An overall comparison of the heuristics versus a baseline can be seen in the figure \ref{fig:compare}

\subsection{Empirical Verification of Fourth-Order Convergence Using the Brusselator}

To validate the effectiveness of the Heuristics found on the reduced scheme \cite{Benner1998}, order convergence studies were performed on each of the heuristics to ensure that each heuristic can provide 4th-order convergence. Each heuristic was applied to an Ordinary Differential Equation (ODE), aka Brusselator, to test the coefficient-based heuristics. The Brusselator is a well-known reaction-diffusion system that exhibits complex, oscillatory behaviour, making it ideal for evaluating accuracy and stability. Its nonlinearity and dynamic nature provide a rigorous test for assessing the order of convergence of the generated schemes. An overview of the 1D-Brusselator equations is given below. 

\begin{equation}
\label{eq:brusselator1}
\frac{dA}{dt} = k_1 - (k_2 + 1)A + A^2B,
\end{equation}
\begin{equation}
\label{eq:brusselator2}
\frac{dB}{dt} = k_2A - A^2B,
\end{equation}

Equations \ref{eq:brusselator1} and \ref{eq:brusselator2} describe the dynamics of the 1D Brusselator system. The equations were integrated from 0 to 50 with a step size of $1 \times 10^{-2}$, where $A$ was set to 1.5 and $B$ set to 3.0, $k_1$ was set to 1.0, and $k_2$ was set to 2.0. This set of coefficients represents the rate at which the chemicals interact with each other over the period. The following graph shows the system’s behavior from the above configurations of the 1D-Brusselator.

\begin{figure}[h]
    \centering
    \includegraphics[scale=0.30]{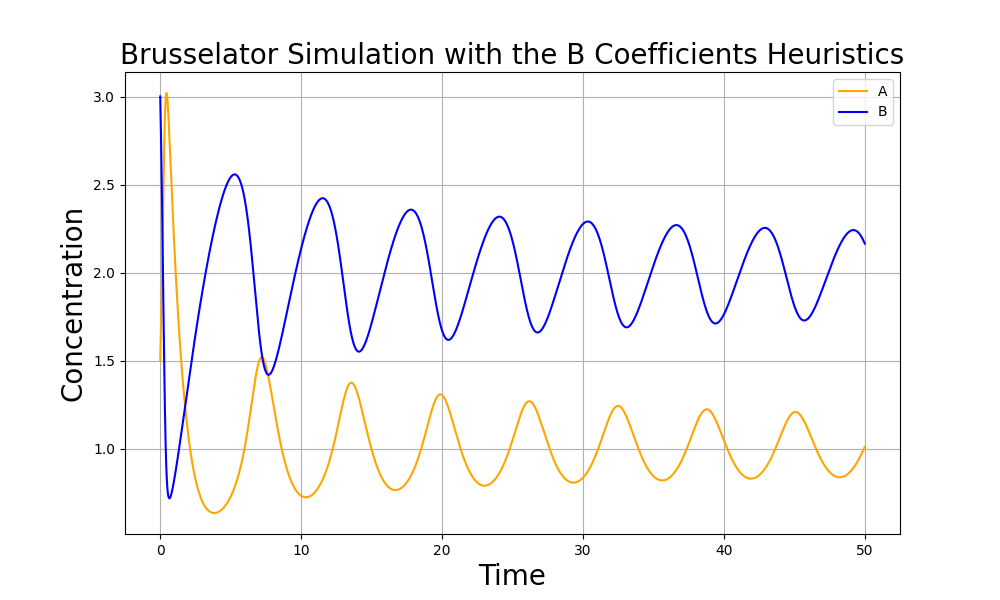}
       \caption{Species A and B concentration dynamics in a 1D Brusselator system over time. The simulation uses heuristic coefficients from the \textbf{a} coefficient table. Species A and B exhibit oscillatory behavior, with A's concentration peaking earlier and lower level than B's. This plot demonstrates the typical oscillatory nature of the Brusselator model.}
    \label{fig:bruss3}
\end{figure}

The 1D-Brusselator system involves two chemicals, “A” and “B”, which undergo a chemical reaction process with a step size of 0.01. The simulation shows species A has a more variable concentration behaviour, as seen in the orange line, which oscillates with a larger amplitude than species B.” On the other hand, species “B” exhibits a smoother, lower-amplitude oscillation, indicating a less dramatic variation in its concentration. These oscillations reflect the inherent nature of the chemical reactions in the system. Each heuristic-based coefficient was tested to ensure that each heuristic produced the correct behaviour of the 1D Brusselator, as seen in the following graph \ref{fig:bruss4}. This behaviour was consistent across all the coefficient heuristics used.

\subsection{2D Brusselator Heuristic Test} 
The following test equation was used with the coefficient-based heuristic schemes the 2D Brusselator \ref{eq:2D-brusselator}. The 2D Brusselator is widely used in numerical analysis due to its ability to model spatial-temporal patterns through reaction-diffusion mechanisms. Its nonlinear dynamics and multidimensional structure make it a non-trivial test case for evaluating the integration scheme's robustness and order of convergence.

\begin{align}\label{eq:2D-brusselator}
\frac{\partial u}{\partial t} &= D_u \left( \frac{\partial^2 u}{\partial x^2} + \frac{\partial^2 u}{\partial y^2} \right) + A - (B + 1) u + u^2 v, \\
\frac{\partial v}{\partial t} &= D_v \left( \frac{\partial^2 v}{\partial x^2} + \frac{\partial^2 v}{\partial y^2} \right) + B u - u^2 v,
\end{align}

The heuristics were used to integrate equation \ref{eq:2D-brusselator}. The following system behaviour graph was produced.

\begin{figure}[h]
    \centering
    \includegraphics[scale=0.45]{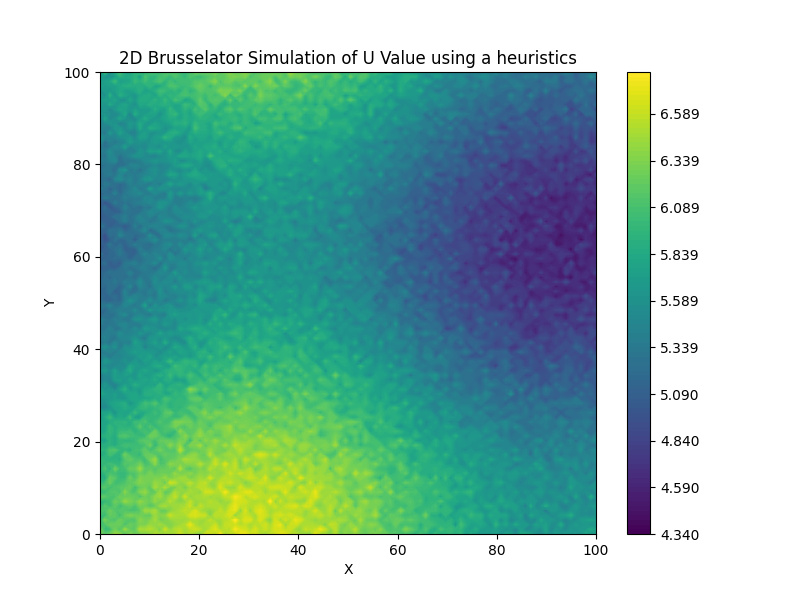}
    \caption{System behaviour of the 2D Brusselator when heuristics are applied. The figure demonstrates how the heuristic-based approach influences the system's dynamics, maintaining stability while achieving accurate solutions. The results illustrate the effectiveness of the heuristics in capturing the underlying oscillatory patterns characteristic of the Brusselator model.}
    \label{fig:bruss4}
\end{figure}

The system behaviour of the following graph \ref{fig:bruss4} is the correct system behaviour of the 2D Brusselator, and this behaviour is consistent across all the heuristics. The configurations of the 2D Brusselator \ref{eq:2D-brusselator} are as follows:

\begin{itemize}
    \item \textbf{Diffusion Coefficients:} Both parameters, \(D_u\) and \(D_v\), were set to 0.1, indicating equal diffusion rates for \(u\) and \(v\). This configuration helps maintain balanced dynamics between the species across the spatial domain.
    
    \item \textbf{Reaction Parameters:} The parameter \(A\) was set to 1.0, which enhances the production of \(u\), while \(B\) was set to 3.0. This configuration influences the production of \(v\) and the consumption of \(u\), leading to complex interactions between the two species and contributing to the system's dynamic behaviour.
    
    \item \textbf{Initial Conditions:} Both \(u\) and \(v\) were set to 0.1. This value sets the starting point for the system’s evolution and allows the simulation to explore the dynamics of \(u\) and \(v\) from a state of initial balance. 
    
    \item \textbf{\(u\) (Species U):} Represents the concentration of a reactant that transforms into product \(v\). The presence and concentration of \(u\) are crucial for initiating and sustaining the reaction process within the system.
    
    \item \textbf{\(v\) (Species V):} Denotes the concentration of the product formed from \(u\). \(v\) plays a dual role by acting as a product and as a catalyst, promoting the production of more \(u\), thereby sustaining the autocatalytic nature of the Brusselator reaction.
\end{itemize}

This enhanced description ensures the reader fully understands the biological or chemical significance of the variables \(u\) and \(v\) within the 2D Brusselator model and their mathematical and dynamic interactions.

These configurations result in a simulation illustrating the interplay between diffusion and reaction processes in the 2D Brusselator. Depending on the precise interaction of the parameters, this often reveals complex patterns such as spirals, spots, or waves.

\subsection{Convergence Analysis of the Coefficient-based Heuristic Schemes}
Convergence analysis is essential for newly created schemes to be usable. The order of convergence of all newly created schemes needs to be of fourth order to ensure the accuracy of the numerical method. The order of convergence of a numerical solution is denoted by the symbol \(p\) and is calculated as follows:

The order of convergence, \( p \), is given by the following formula:
\begin{equation} \label{eq:order_of_convergence}
p = \frac{\log\left(\frac{e_1}{e_2}\right)}{\log\left(\frac{h_1}{h_2}\right)}
\end{equation}
where \( e_1 \) and \( e_2 \) are the errors for step sizes \( h_1 \) and \( h_2 \), respectively.

The order of convergence was calculated for the 1D Brusselator by varying the step sizes from 0.1 to 0.001 over twenty evenly spaced intervals. The error was calculated using the \textit{L2} norm, and the order of convergence was determined using the equation \ref{eq:order_of_convergence}, where \( e_1 \) and \( e_2\) represent the errors associated with the step sizes \( h_1 \) and \( h_2\), respectively:

\begin{equation}
\label{eq:order_equation}
\mathrm{Order} = \frac{\log(e_1 / e_2)}{\log(h_1 / h_2)}
\end{equation}

In equation \ref{eq:order_equation}, $h1$ is the initial step size (e.g., 0.1), and $h2$ is the final step size (e.g., 0.001) used for discretization in numerical simulations, which is calculated using the \textit{L2} norm given by the equation \ref{eq:l2}.

\begin{equation}\label{eq:l2}
L_2 \text{ Error} = \sqrt{\frac{1}{N} \sum_{i=1}^N \| y_{\text{num},i} - y_{\text{ref},i} \|^2}
\end{equation}

Each set of heuristics was evaluated using both 1D and 2D Brusselator test cases to verify that every generated Butcher tableau achieves fourth-order convergence. The accompanying graphs confirm that all heuristic-driven tableaus meet the convergence criteria, demonstrating that the proposed methodology consistently produces robust schemes even over more than 100 runs of the interior point optimizer..

\begin{figure}[h]
    \centering
    \includegraphics[scale=0.50]{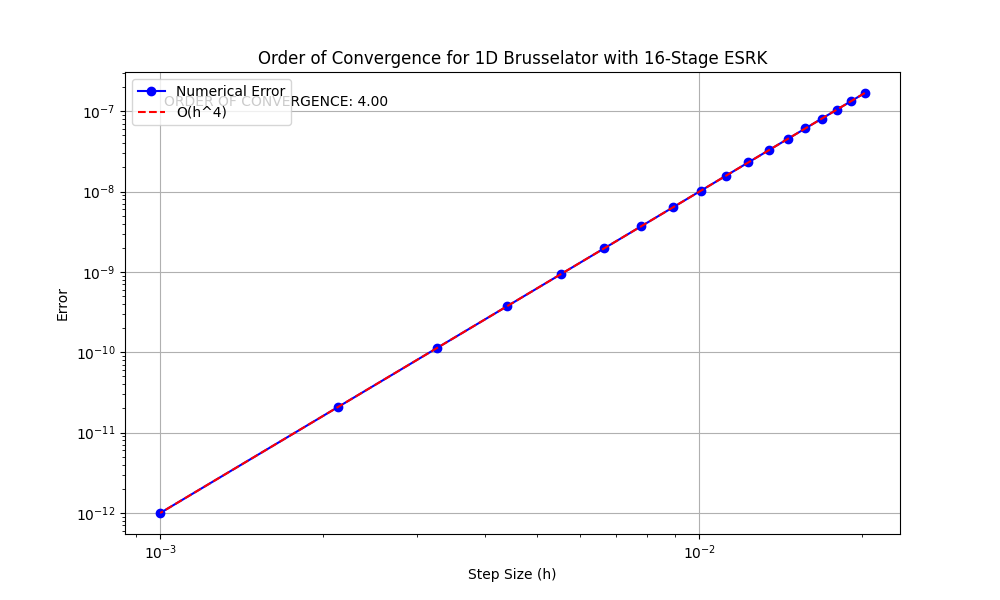}
    \caption{1D Brusselator order of convergence with the \textbf{a} coefficient heuristics being used. When the heuristic Butcher tableaus are used, they provide fourth-order convergence, which is the correct order of convergence needed to produce accurate solutions.}
    \label{fig:order_1}
\end{figure}

Each heuristic was tested from the order convergence studies tested on the 1D Brusselator equation \ref{fig:order_1}, and all produced fourth-order convergence on this test problem. Next, order convergence studies were performed on the 2D Brusselator equation \ref{eq:2D-brusselator}.

\begin{figure}[h]
    \centering
    \includegraphics[scale=0.55]{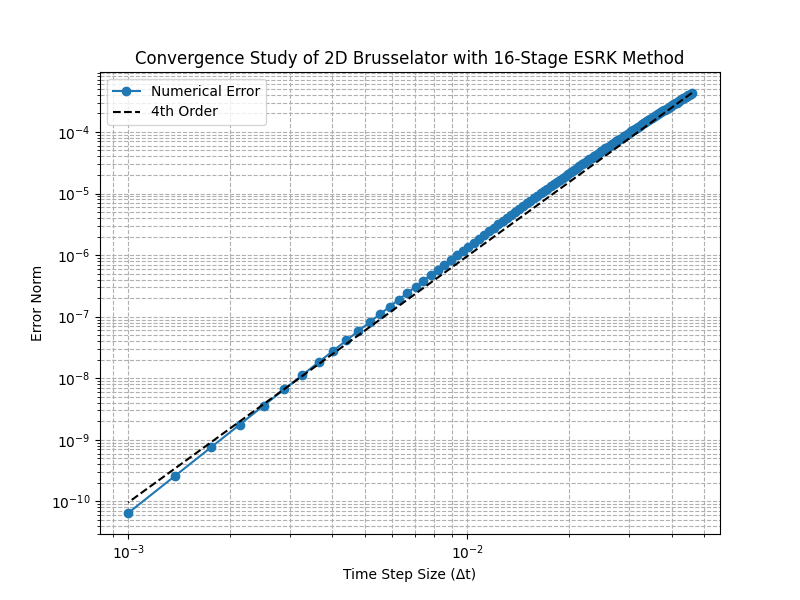}
    \caption{2D Brusselator order of convergence with studies using heuristics. This shows the heuristics can produce 4th-order convergence on a 2D Brusselator. The results validate the effectiveness of the heuristic approach in achieving high-order accuracy while maintaining stability across iterations. The observed convergence behaviour aligns well with theoretical predictions.}
    \label{fig:2D_A}
\end{figure}

From the order convergence studies of the 2D Brusselator equation \ref {fig:2D_A}, each created heuristic provided 4th-order convergence, which is needed for the schemes to be usable in real-world applications.

\subsection{Order Convergence Studies with the Naive Stokes Equations}

We extended the convergence studies to the steady-state form of the Stokes equations to further validate the effectiveness of the heuristics found in the reduced scheme. These equations, often called the "Naive Stokes equations," provide a simplified yet rigorous test for numerical methods used in fluid dynamics. Their structure is particularly well-suited to examine the stability and accuracy of the heuristics developed in this work. 

The Naiver-Stokes equations describe the motion of a slow-moving, incompressible fluid and are given as follows:

\begin{align}
-\nabla p + \nu \nabla^2 \mathbf{u} &= \mathbf{f}, \label{eq:momentum} \\
\nabla \cdot \mathbf{u} &= 0, \label{eq:continuity}
\end{align}

where \( \mathbf{u} = (u_x, u_y) \) is the velocity vector field, \( p \) is the pressure, \( \nu \) is the kinematic viscosity, and \( \mathbf{f} = (f_x, f_y) \) is the external force. The momentum equation \eqref{eq:momentum} governs the balance of forces in the fluid, while the continuity equation \eqref{eq:continuity} ensures incompressibility.

\subsubsection*{Order of Convergence}

The order of convergence for the reduced scheme was evaluated by computing the errors in the velocity field against a reference solution. Errors were measured using the \( L_2 \)-norm and calculation \ref{eq:l2} the following order convergence graphs were produced.

\begin{figure}[h]
    \centering
    \includegraphics[scale=0.50]{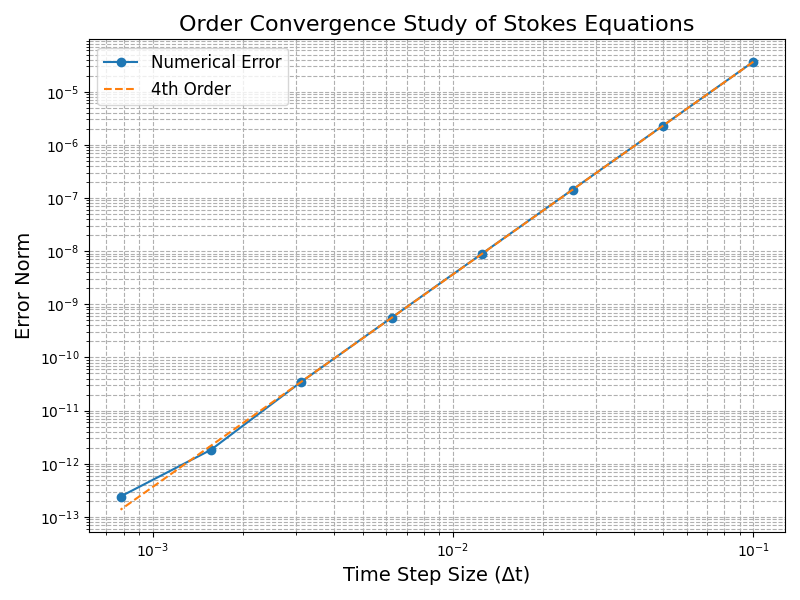}
    \caption{Order convergence study for the Stokes equations using the 16-stage 4th-order ESRK method. The numerical error (blue points) demonstrates a consistent 4th-order convergence rate, as indicated by the slope of the reference line (orange dashed line). The results validate the accuracy and stability of the reduced scheme applied to the Naive Stokes equations.}
    \label{fig:order_5}
\end{figure}
\subsection{Discussion of the Results}\label{subsection:discussion}

The results of this study demonstrate the significant potential of applying heuristic-driven optimization to enhance the efficiency of reduced Runge-Kutta schemes, specifically our 16-stage, 4th-order ESRK scheme. As detailed in Table \ref{tab:baseline_stats}, the baseline performance—without any heuristics—averaged 2010 iterations over 100 runs. By integrating heuristics into the parameter-reduced framework, we achieved notable improvements in convergence efficiency.

\subsection{Baseline Performance}\label{section:dicussion}

Our baseline evaluation (see Table \ref{tab:baseline_stats}) shows that the reduced scheme, when executed without heuristic guidance, requires an average of 2010 IPOPT iterations to converge. This value serves as a benchmark for assessing the impact of our heuristic enhancements.

\subsection{Heuristic Performance}

Incorporating heuristics into the optimisation process resulted in a substantial performance boost. The best-performing heuristic, \( b_5 = a_{2,1}^2 \), reduced the average runtime to 1501 iterations over 100 runs, corresponding to an approximately 25\% decrease in IPOPT iterations (refer to Table \ref{table:Reduce_performance}). Figure \ref{fig:compare} further illustrates the performance improvements by comparing the baseline and heuristic-enhanced run times via a box plot.

These results validate the efficacy of our heuristic approach, as summarised in Table \ref{tab:key_findings}. Importantly, despite the significant reduction in iterations, the convergence studies (Figures \ref{fig:order_1}, \ref{fig:2D_A}, and \ref{fig:order_5}) confirm that the fourth-order accuracy of the scheme is maintained. Overall, the integration of heuristics into our reduced scheme not only enhances computational efficiency but also preserves the high-order accuracy required for robust, high-fidelity simulations.

\subsection{Implications of the Results}

The outcomes of our study have several significant implications, as summarised in Table \ref{tab:implications}. In particular, the reduction in IPOPT iterations from 2,010 to 1,501 demonstrates notable efficiency gains, which can lead to substantial savings in computation time and energy for large-scale simulations. The successful application of our heuristic approach validates the hypothesis that intelligently designed heuristics can exploit the reduced parameter space effectively, without compromising the scheme's stability or fourth-order accuracy. Moreover, while a 25\% reduction in runtime is already significant given the optimised, low-storage nature of the scheme, there remains potential for further improvements as additional heuristics are explored. Importantly, the convergence studies on the 1D Brusselator, 2D Brusselator, and Naive Stokes equations confirm that the introduction of heuristics does not degrade the high-order accuracy of the scheme. These findings collectively underscore the robustness of our methodology and its promise for enhancing computational efficiency in high-fidelity simulations.

\begin{table}[h!]
\centering
\caption{Implications of the Results of using heuristic}
\label{tab:implications}
\begin{tabular}{|p{4cm}|p{8cm}|}
\hline
\textbf{Implication} & \textbf{Description} \\
\hline
Efficiency Gains & The reduction from 2,010 to 1,501 iterations in IPOPT run times demonstrates a significant improvement in computational efficiency. This reduction can translate into notable resource savings in computation time and energy consumption for large-scale applications or high-fidelity simulations. \\
\hline
Validation of the Heuristic Approach & The success of the best heuristic validates the practical value of heuristic-based optimisations, confirming that intelligently designed heuristics can effectively exploit the reduced parameter space without compromising the scheme's stability or order of accuracy. \\
\hline
Potential for Further Improvements & While the best heuristic achieved a 25\% runtime reduction, the framework developed in this study leaves room for additional enhancements. Further heuristics could yield even more significant efficiency gains as computational demands increase. \\
\hline
Maintained 4th-order Convergence & Crucially, the introduction of heuristics did not degrade the 4th-order accuracy of the reduced scheme. Convergence studies on test problems (including the 1D Brusselator, 2D Brusselator, and Stokes equations) confirm that the approach maintains robust, high-order accuracy. \\
\hline
\end{tabular}
\end{table}

\subsection{Broader Impacts}

The findings of this study extend beyond the immediate improvements in computational efficiency and accuracy. The demonstrated enhancements render reduced schemes with heuristics a practical and compelling solution for tackling complex, real-world problems in numerical analysis and computational fluid dynamics. Moreover, the methodologies developed here establish a solid foundation for future research in heuristic-guided optimisation. By building upon this work, subsequent studies can further advance low-storage numerical methods, ultimately leading to even more efficient and scalable algorithms for high-fidelity simulations.
\section{Conclusions and Future Work}\label{section:conclusions}

This study underscores the effectiveness of combining heuristic-driven approaches with reduced schemes to achieve significant efficiency gains in computational performance. The best-performing heuristic demonstrated a 25\% reduction in average runtime while maintaining high-order accuracy, highlighting the potential of such techniques to enhance efficiency and reliability in solving computational problems. These findings contribute to the growing research on reduced schemes and emphasise their practical utility across diverse computational applications.

Several promising avenues exist to refine and expand this work. First, exploring heuristic designs that extend beyond a fixed length of three variables could lead to more optimised solutions with improved adaptability. Incorporating variable-length heuristics or conducting a more exhaustive exploration of reduced schemes may uncover heuristics with greater computational efficiency. Employing advanced search techniques, such as adaptive or prioritised algorithms, could streamline this process by focusing computational resources on the most promising regions of the search space. Moreover, testing the proposed heuristics across a broader range of problems, including stiff ordinary differential equations (ODEs) and other challenging computational benchmarks, would enhance their robustness and generalizability. Techniques like dimensionality reduction or dynamically adjusting the heuristic length could also help balance the computational cost and optimise potential trade-offs.

For future works, several areas of further investigation could be explored. One possibility is integrating machine learning techniques to automatically generate and adapt heuristics based on the specific characteristics of the problem. Additionally, utilising parallel and distributed computing frameworks could facilitate the evaluation of heuristics over more extensive and more complex search spaces. Another potential direction is the application of these techniques in real-time systems, where computational efficiency is critical in fields like robotics, finance, and dynamic system modelling. Expanding the study to include multi-objective optimisation problems could also provide a more comprehensive assessment of heuristic-driven approaches by addressing multiple performance criteria simultaneously. These directions could help refine heuristic-based methods and broaden their applicability in various domains.

\bibliographystyle{plainnat}  

\end{document}